\documentclass[lettersize,journal]{IEEEtran}
\usepackage{amsmath,amsfonts}
\usepackage{algorithmic}
\usepackage{algorithm}
\usepackage{array}
\usepackage[caption=false,font=normalsize,labelfont=sf,textfont=sf]{subfig}
\usepackage{textcomp}
\usepackage{stfloats}
\usepackage{url}
\usepackage{verbatim}
\usepackage{graphicx}
\usepackage{cite}
\usepackage[dvipsnames]{xcolor}
\usepackage{booktabs}
\usepackage{multirow}

\begin{document}

\title{Disambiguating Monocular Reconstruction of 3D Clothed Human with Spatial-Temporal Transformer}

\author{Yong Deng, Baoxing Li, Xu Zhao*,~\IEEEmembership{Member,~IEEE,}
\thanks{This paper was produced by the IEEE Publication Technology Group. They are in Piscataway, NJ.}
\thanks{Manuscript received April 19, 2021; revised August 16, 2021.}}

\markboth{Pattern Recognition}%
{Shell \MakeLowercase{\textit{et al.}}: A Sample Article Using IEEEtran.cls for IEEE Journals}


\maketitle
\begin{abstract}
Reconstructing 3D clothed humans from monocular camera data is highly challenging due to viewpoint limitations and image ambiguity. While implicit function-based approaches, combined with prior knowledge from parametric models, have made significant progress, there are still two notable problems. Firstly, the back details of human models are ambiguous due to viewpoint invisibility. The quality of the back details depends on the back normal map predicted by a convolutional neural network (CNN). However, the CNN lacks global information awareness for comprehending the back texture, resulting in excessively smooth back details. Secondly, a single image suffers from local ambiguity due to lighting conditions and body movement. However, implicit functions are highly sensitive to pixel variations in ambiguous regions. To address these ambiguities, we propose the Spatial-Temporal Transformer (STT) network for 3D clothed human reconstruction. A spatial transformer is employed to extract global information for normal map prediction. The establishment of global correlations facilitates the network in comprehending the holistic texture and shape of the human body. Simultaneously, to compensate for local ambiguity in images, a temporal transformer is utilized to extract temporal features from adjacent frames. The incorporation of temporal features can enhance the accuracy of input features in implicit networks. Furthermore, to obtain more accurate temporal features, joint tokens are employed to establish local correspondences between frames. Experimental results on the Adobe and MonoPerfCap datasets have shown that our method outperforms state-of-the-art methods and maintains robust generalization even under low-light outdoor conditions.
\end{abstract}


\begin{figure}[!t]
\centering
\includegraphics[width=\linewidth]{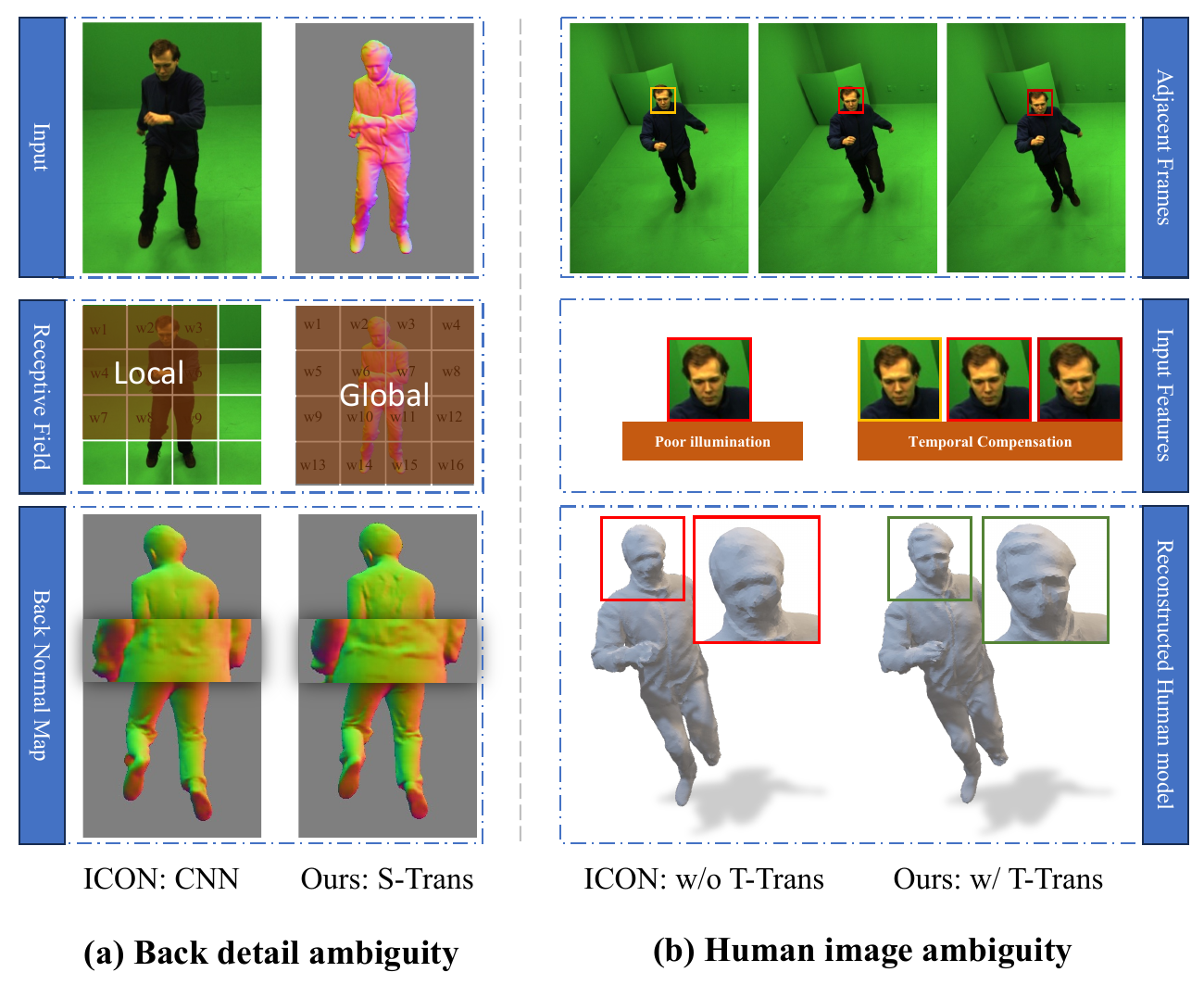}
\caption{The current method of implicit reconstruction has two problems: (a) Ambiguity in back details due to the invisibility. The perception of global information can help the normal network to infer back details. (b) Ambiguity in human images due to lighting conditions and body movement. However, there may be some improvement in corresponding areas in adjacent frames.}
\label{problem}
\end{figure}

\section{Introduction}
\IEEEPARstart{T}{hree-dimensional} human surface reconstruction aims to digitize detailed human surfaces, including hair, clothing, and other features. It finds extensive applications in areas such as film and television production, gaming and virtual reality. Existing methods reconstruct the 3D human model from data captured by a RGBD camera, Inertial Measurement Units (IMUs), or multiple sensors. In comparison, reconstructing the 3D human from monocular RGB camera is more economical and convenient. The video data captured by a monocular camera provides rich multimedia information in spatial and temporal dimensions to help reconstruct the human body's posture, shape, and dynamic features.

The prevailing methodologies in the field can be broadly classified into two categories: model-based methods and implicit function-based methods. Model-based methods \cite{Tex2Shape,Octopus,HPSD,HPSCG} represent clothed humans using parametric models combined with vertex offsets. Due to limitations in the topology of the parametric model, these methods have limited ability to capture intricate details. Implicit function-based methods \cite{PIFu,ARCH,ARCH++} reconstruct human models by iteratively querying voxel points from a predefined 3D volume through an implicit network, theoretically promising infinite resolution. Despite the capacity of implicit functions to encapsulate all human details, the absence of explicit human priors may lead to unreasonable geometric shapes. In response to this limitation, \cite{PaMIR} combines the human priors of parametric models with the representational ability of implicit functions. Three-dimensional features are extracted from the parametric model and fed into the implicit network. However, the invisible detail cannot be directly captured from images, and implicit networks struggle to learn the intricate features of the back details. Consequently, the reconstruction of the back region tends to exhibit smoothness. To address this, some methods \cite{ICON,PIFuHD,IntegratedPIFu} employ normal maps as a middle agent, delegating the prediction of back details to a specialized normal prediction network. The implicit network then focuses on reconstructing human details based on the information of normal maps.

The above methods have produced reliable reconstructions, but two problems remain: {\emph{1) Back detail ambiguity.}} The details of back normal maps are ambiguous due to their invisibility. Current methods \cite{ICON,PIFuHD,IntegratedPIFu} employ residual convolutional networks \cite{pix2pix} to predict back normal maps. However, each neuron in the network's convolutional layers concentrates solely on the local region of input data. The absence of global information hinders the network's ability to learn precise back texture, leading to a smooth reconstruction of the human back as shown in Fig.\ref{problem}. {\emph{2) Human image ambiguity.}} The implicit network takes pixel-aligned features extracted from the images as input, making it highly sensitive to variations in pixel values. When the captured images exhibit local ambiguity due to lighting conditions or body movement, the reconstruction result may easily generate errors in local details. Current methods struggle to accurately determine human body details when faced with both types of ambiguity.

To address the aforementioned issues, we propose a {\bf{S}}patial {\bf{T}}emporal {\bf{T}}ransformer network for 3D clothed human reconstruction (STT). To address the ambiguity of back details, the extraction of global information is crucial. The self-attention mechanism facilitates the establishment of global correlations across different regions of an image. Therefore, a spatial transformer (S-Trans) is introduced for normal map prediction, which can extract global information for normal inference in invisible regions. Meanwhile, the image information from adjacent frames in a temporal sequence can compensate for local ambiguous areas in human images. To extract useful temporal information from temporal segments and enhance the reconstruction of the current frame, we propose a temporal transformer (T-Trans). It utilizes a multi-head self-attention mechanism to integrate information from adjacent frames, extracting temporal features for enhancing the quality of implicit reconstruction.

Fig. \ref{pipeline} illustrates the pipeline of our method. Taking a clip of images as input, the spatial transformer initially predicts the normal features. Two spatial transformers with identical structures are employed to separately predict the front and back normal maps. The front-normal S-Trans uses the image as input, while the back-normal S-Trans utilizes the front normal map. Subsequently, the temporal transformer extracts temporal features form the predicted sequence of normal maps. To facilitate the network in capturing internal correlations between frames, joint tokens are introduced as guidance in the learning process. Finally, the implicit function takes the 2D features from normal maps, 3D features from parametric models and temporal features as input to generate the ultimate detailed human model.

The main contributions of our work are threefold:
\begin{itemize}
\item{A spatial transformer is employed to learn global correlations, enhancing the quality of predicted normal maps and alleviating the over-smoothing issue caused by ambiguity in back details.}
\item{A temporal transformer guided by joint positions is proposed to capture temporal features. These features mitigate the effects of local ambiguity in human images caused by lighting conditions and body movement.}
\item{As far as we know, It is the first attempt to apply spatial-temporal transformer into 3D clothed human reconstruction. It achieves state-of-the-art results on the Adobe \cite{Adobe} and MonoPerfCap \cite{MonoPerfCap} datasets.
}
\end{itemize}

\section{Related Work}

Current approaches to monocular human modelling focus on monocular images (Sec. \ref{mono_image}). Monocular video-based methods (Sec. \ref{mono_video}) mainly study real-time reconstruction and avatar generation without considering the temporal information. In most methods, there is a consensus that the normal map (Sec. \ref{normal}) is an important intermediate representation for monocular human modelling, which plays an important role in supporting texture understanding.

\subsection{Human Modelling with Monocular Image}
\label{mono_image}

The methods for 3D Clothed Human Reconstruction are primarily categorized into explicit reconstruction and implicit reconstruction. Explicit reconstruction methods \cite{Tex2Shape,Octopus,HPSD,HPSCG,CAPE,HMD,IUVD} represent clothing details by using parametric models \cite{SMPL,SMPL-X,SMPLify} combined with vertex offsets. The advantage of such methods lies in their ability to maintain the integrity of the human body structure in challenging situations with the assistance of prior information from the parametric model. However, due to the limitations in the topological structure of the parametric model, the accuracy of reconstruction is constrained.

Saito et al. \cite{PIFu} introduce Pixel-aligned Implicit Functions (PIFu). PIFu extracts pixel-aligned features fed into an implicit network represented by multi-layer perceptrons (MLPs), and obtains the reconstruction result through a point-wise query approach. The recovery of a 3D human from 2D features presents a fundamentally ill-posed problem. Additionally, implicit reconstruction methods, lacking human priors, may lead to the reconstruction of unreasonable structures. Therefore, PaMIR \cite{PaMIR} proposes the integration of parametric models into implicit networks to extract voxel-aligned 3D features. Unlike PaMIR, TransPIFu \cite{TransPIFu} employs a transformer to extract 3D voxel features. The self-attention mechanism effectively learns 3D geometric correlations from 2D image features. Implicit reconstruction methods heavily rely on point features extracted from the image, making them particularly sensitive to changes in local pixel values. Errors in reconstruction can arise from local image ambiguity caused by factors such as lighting or body movement. Therefore, we propose a temporal transformer to extract temporal features, enhancing the reconstruction of locally ambiguous regions in the current frame by utilizing information from adjacent frames.

\begin{figure*}[!t]
\centering
\includegraphics[width=\linewidth]{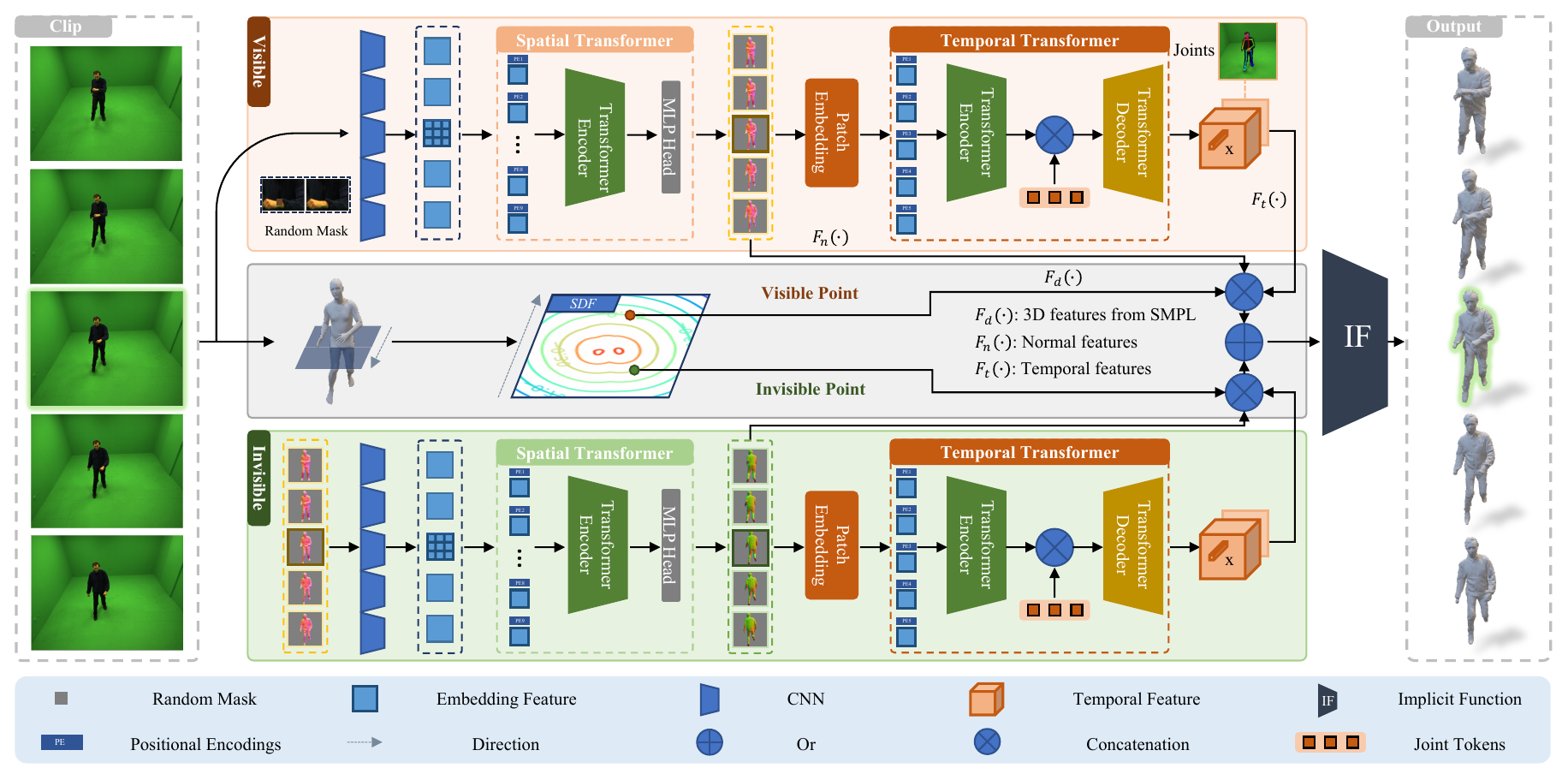}
\caption{Our method comprises two pivotal modules: (1) a {\bf{Spatial Transformer}} (S-Trans) for normal map prediction and (2) a {\bf{Temporal Transformer}} (T-Trans) for temporal information extraction from sequences of normal maps. Different procedures are employed for visible (\textcolor{orange}{orange box}) and invisible (\textcolor{green}{green box}) point, with the primary distinction lying in the normal map prediction module. Two S-Trans with identical structures are employed to separately predict the front and back maps. The front-normal S-Trans uses the image as input, while the back-normal S-Trans utilizes the front normal map. The T-Trans utilizes the same module across both prediction processes. The Joint Tokens are employed to guide the correspondence between adjacent frames in network learning. The output joint positions serve as a supervisory signal for the network. In addition, to enhance the network's learning in ambiguous areas, we introduced random mask during the training process in the second stage. Finally, the input of the implicit function consists of 2D features (Normal map), 3D features (SDF), and temporal features (T-Trans).}
\label{pipeline}
\end{figure*}

\subsection{Human Modelling with Monocular Video}
\label{mono_video}

When reconstructing a 3D clothed human body from data captured by a single RGB camera, it is necessary not only to independently consider the reconstruction of each frame image but also to take into account the reconstruction speed and the utilization of temporal information.
MonoPort \cite{MonoPort} proposes a hierarchical surface localization algorithm and direct rendering method to accelerate the speed of reconstruction from videos. FOF \cite{FOF} proposes a new representation, Fourier Occupancy Field, which represents 3D geometry as a 2D map, improving reconstruction quality and speed.
However, this method lacks the utilization of temporal information. To harness information from each video frame and reconstruct higher-quality human models, \cite{Selfrecon,Vid2avatar,Instantavatar,HF-Avatar} train an avatar from video clips and then transform it into the pose of the corresponding frame to obtain the reconstruction results. While these methods enhance the accuracy of reconstruction by incorporating information from each frame in the video, they require retraining for different reconstructed objects. In contrast, TCR \cite{TCR} optimizes reconstruction results by preserving human shape information from past frames using a canonical model. For each frame, model reconstruction merely requires blending the canonical model with the current reconstruction, resulting in an optimized reconstruction output. Although TCR leverages optimization steps to enhance reconstruction accuracy, it still relies on the accuracy of single-frame reconstruction. To leverage temporal information during single-frame reconstruction and enhance its accuracy, our method employs transformers to establish temporal correlations between adjacent frames in the video and extracts valuable temporal information to augment the reconstruction quality of the current frame.

\subsection{Normal Prediction}
\label{normal}

The prediction of details in the invisible region is an ill-posed problem as it cannot be directly observed from images. The inherent ambiguity of back details makes it challenging to directly estimate back details from images.  \cite{ICON,ECON,PIFuHD,IntegratedPIFu} utilize the normal map as a proxy for human geometry, transferring the prediction of back details to a normal prediction network. These methods allow MLPs to focus on recovering details from the normal maps. PIFuHD \cite{PIFuHD} employs a residual convolutional network \cite{pix2pix} to infer normals from images. Similarly, ICON \cite{ICON} uses the same network for normal prediction but incorporates the parametric model's normal map as priors into the normal prediction network. In addition, ECON \cite{ECON} utilizes MRF loss \cite{MRF} to propagate local and global information obtained from images in order to enhance the prediction of back normals. However, convolutional networks struggle with long-range dependencies, resulting in insufficient global understanding. The lack of global information leads to excessively smooth back normal predictions. Therefore, we introduce a spatial transformer that leverages attention mechanisms to establish global awareness, enhancing back details prediction and overall reconstruction quality.

\begin{figure*}[!t]
\centering
\includegraphics[width=\linewidth]{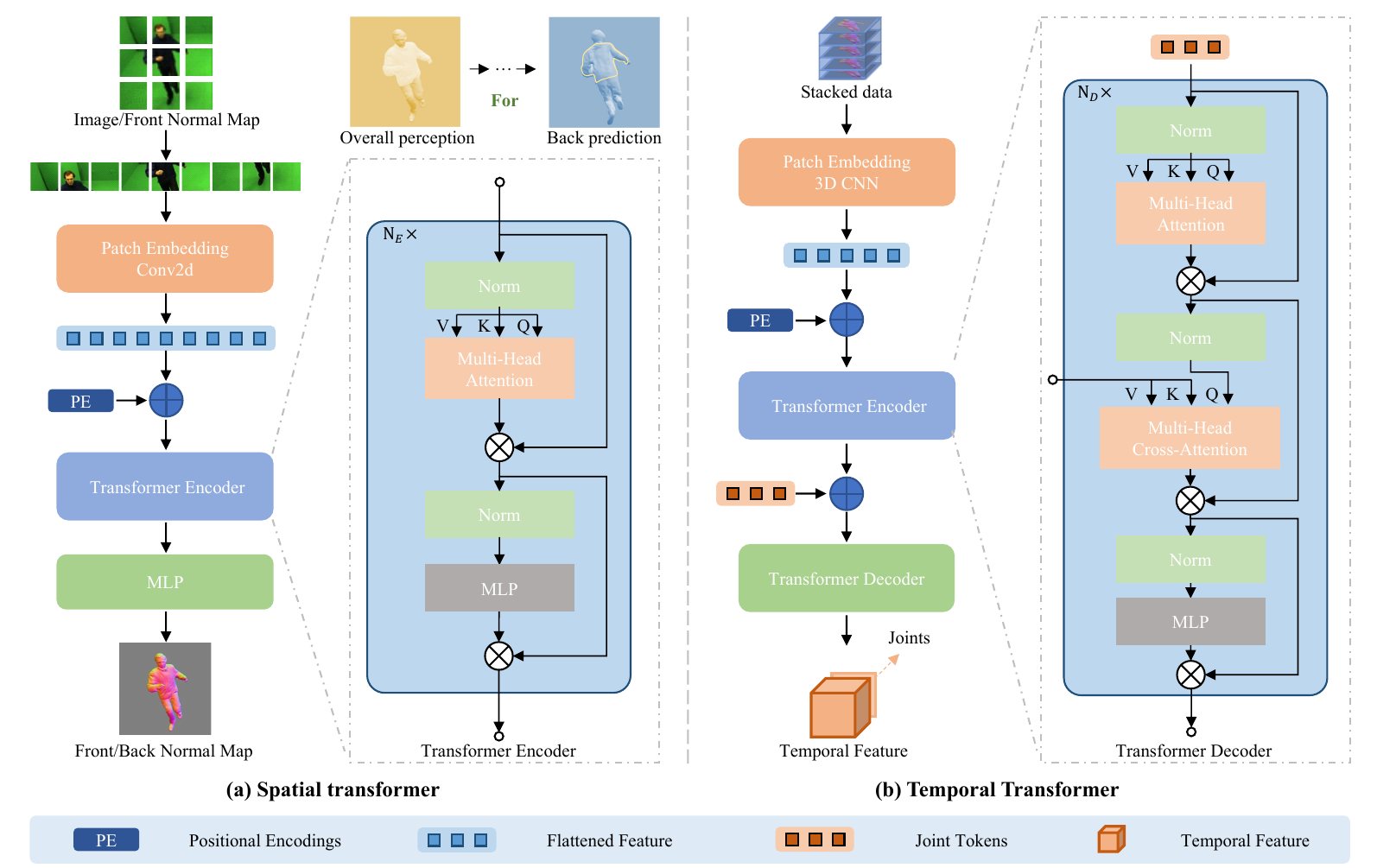}
\label{method}
\caption{The internal architecture of the transformer module. (a) shows the spatial transformer employed for normal prediction, and (b) illustrates the temporal transformer utilized for temporal feature extraction. The structures of the two encoders are identical, but their parameter settings are different. The spatial transformer uses a multi-layer perceptron to transform the encoded global features into a normal map. The temporal transformer uses a decoder to aggregate the features between frames and convert them into temporal features. Additionally, joint tokens are introduced in the temporal transformer. The output joint positions serve as supervised guidance to assist the network in learning the correspondence between frames.}
\label{method}
\end{figure*}

\section{Method}
\subsection{Method Overview}
Our method shares a similar architecture with ICON \cite{ICON}, as illustrated in Fig. \ref{pipeline}. We also employ normal features as inputs to the implicit function. However, we replace the original 2D CNN with a Spatial Transformer (Sec. \ref{SpatialT}) to enhance the prediction of back normal maps. Additionally, we adopt a Temporal Transformer (Sec. \ref{TemporalT}) to extract temporal features for enhancing predictions in ambiguous regions of images. To achieve superior results, we utilize the post-optimization steps (Sec. \ref{MeshR}) proposed in our earlier work on TCR \cite{TCR}.

For each frame, we initially employ PyMAF \cite{PyMAF} to obtain the corresponding SMPL \cite{SMPL} mesh $\mathcal{M}(\theta,\beta)\in \mathbb{R}^{N\times 3}  $, where $\theta$ and $\beta$ represent the pose and shape parameters respectively. Simultaneously, a spatial transformer is used to estimate the ``front'' (visible side) and ``back'' (invisible side) normal maps $\mathcal{N}=\{\mathcal{N}_f,\mathcal{N}_b\}$. Subsequently, a temporal transformer derives temporal features $\mathcal{F}_t=\{\mathcal{F}_t^f,\mathcal{F}_t^b\}$ based on the normal maps from adjacent frames. The input features for the implicit network are then: 
\begin{equation}
\label{feature}
F_V=[F_d(V),F_n(V),F_t(V)],
\end{equation}
where $F_d$ is the signed distance from a query voxel $V$ to the closet point on SMPL mesh $V^c\in \mathcal{M}$, $F_n$ is the Corresponding normal vector obtained from $\mathcal{N}$ and $F_t$ is the feature vector from temporal features $\mathcal{F}_t$. The visibility of voxel $V$ depends on the visibility of $V^c$. $V^c$ is visible when the angle between the line of sight and the surface normal is less than 90 degrees. Otherwise, it is invisible. Finally, the implicit function $f(V)$ can be expressed as:
\begin{equation}
\label{implicit}
f(V)=g(F_V,Z),
\end{equation}
where Z is the depth of voxel $V$ along the ray and $g$ is an implicit network represented by Multi-layer Perceptrons (MLPs). All occupied voxels can be extracted by applying a predefined threshold of 0.5. Subsequently, the marching cubes algorithm \cite{MarchingC} is utilized to generate final human mesh. Furthermore, several optimization processes are employed to refine the human model.

\subsection{Spatial Transformer}
\label{SpatialT}
Previous methods for normal prediction employed a CNN. The convolutional operations primarily focused on local receptive fields, achieving commendable results in recovering local details of front normal maps. However, due to the lack of global information, the network struggled to capture the overall correlation of normals, resulting in limited effectiveness in predicting back normal maps. Self-Attention mechanism \cite{Attention} has the capability to establish correlations between different positions in space. We employ spatial transformers for the normal prediction.

As shown in Fig. \ref{method}, when a single image $x\in\mathbb{R}^{H\times W\times C}$ is input, it is initially divided into $N$ patches, where $N=HW/P^2$ and $P$ is patch size. Subsequently, a 2D convolution operation $E$ maps them into D-dimensional vectors as patch embeddings. These patch embeddings are then concatenated with positional embeddings $E_p\in\mathbb{R}^{N\times D}$ and then the input of encoder is:
\begin{equation}
\label{z_0}
z_0=[x_p^1E;x_p^2E;...;x_p^N]+E_p.
\end{equation}
The Transformer encoder \cite{VIT} comprises Multi-head Attention (MHA) and Multi-layer Perceptron (MLP) blocks. LayerNorm (LN) and residual connections are applied before and after each block, respectively. If the number of encoder layers is denoted as 
$L$, the output of the encoder is:
\begin{align}
z_l^{'}&=MHA(LN(z_{l-1}))+ z_{l-1},&l=1...L \\
z_l&=MLP(LN(z_l^{'}))+ z_l^{'},&l=1...L
\end{align}
Ultimately, we employ a Multi-Layer Perceptron (MLP) to project the output dimensions of the transformer encoder onto the dimensions of the original image and then the normal map $\mathcal{N}$ can be obtained:
\begin{equation}
\mathcal{N}=MLP(z_l)\in \mathbb{R}^{H\times W\times C} .
\end{equation}

\noindent {\bf{Model training.}} Spatial transformers are trained independently, and their parameters are subsequently fixed to train jointly with other modules. Two spatial transformers, with the same structure, are employed to predict front and back normal maps separately. Similarly to TCR \cite{TCR}, for back normal map prediction, we use the front normal map as input. The loss function employed in our training is as follows:
\begin{equation}
\label{loss}
\mathcal{L}_N=\mathcal{L}_{MSE}+\lambda_{PERC}\mathcal{L}_{PERC}.
\end{equation}
$\mathcal{L}_{MSE}$ is Mean Squared Error (MSE) of normal maps:
\begin{equation}
\label{loss}
\mathcal{L}_{MSE}=\frac{1}{H \times W} \sum_{i=1}^{H} \sum_{j=1}^{W} \|\mathcal{N}_{\text{gt}}(i,j) - \mathcal{N}_{\text{pred}}(i,j)\|^2,
\end{equation}
where $\mathcal{N}_{\text{gt}}$ is generated by rendering from ground-truth clothed human model. $\mathcal{L}_{PERC}$ is perceptual loss \cite{PERC}, employed to quantify the high-level perceptual and semantic differences between images. $\lambda_{PERC}$ is hyper-parameter.

\subsection{Temporal Transformer}
\label{TemporalT}
Single-frame images often exhibit ambiguous regions, and implicit reconstruction is highly sensitive to variations in pixel values, leading to potential errors in 3D geometric reconstruction. These ambiguous regions can be improved through corresponding information from adjacent frames. Therefore, we utilize a temporal transformer to extract temporal information, aiming to improve the inaccuracies in features caused by the ambiguity of human images.

As illustrated in the Fig. \ref{method}, we utilize the predicted normal map sequence as the input to the spatial transformer. A clip with $T$ frame is denoted by $\mathcal{N}_{clip}\in \mathbb{R}^{T\times H\times W\times C}$. 
To capture spatio-temporal features, we initially employ a 3D CNN to generate the fusion feature $\mathcal{F}_c\in \mathbb{R}^{T\times H\times W\times D_c}$. A feature sequence is obtained by flattening temporal dimensions of $\mathcal{F}_c$. Subsequently, we integrate it with learnable positional embeddings and feed them into a transformer encoder, sharing a structure identical to the encoder in the spatial transformer. Given a normal map sequence, the transformer encoder produces aggregated spatiotemporal features $\mathcal{F}_e\in \mathbb{R}^{T\times HW\times D_e}$. 

\noindent {\bf{Joint Tokens.}} The transformer decoder aims to map high-level spatiotemporal features $\mathcal{F}_e$ from the encoder to instance-level temporal features $\mathcal{F}_t\in \mathbb{R}^{H\times W\times D_t}$. To enable the network to learn human body correspondence across frames, the Transformer decoder incorporates joint tokens $\mathcal{J}=\{j_1,j_2,...,j_K\}$ to regress the joint position of each frame. With the spatiotemporal features $\mathcal{F}_e$ and joint tokens $\mathcal{J}$, the transformer decoder produces joint features $\mathcal{F}_j\in \mathbb{R}^{H\times W\times D_j}$ and temporal features $\mathcal{F}_t$ using self-attention and cross-attention blocks. Then, pixel-aligned temporal feature can be obtained through bilinear sampling $\mathcal{T}$ from the feature $\mathcal{F}_t$:
\begin{equation}
F_t(V)=\mathcal{T}(\mathcal{F}_t,\pi (V)) .
\end{equation}
where $\pi (V)$ denotes the 2D projection of the 3D voxel $V$.

\noindent {\bf{Model training.}} The temporal transformer and the implicit network are both trained during the second stage. To enhance the network's inference capability in blurry regions, we employ a random mask strategy. We randomly sample an $n\times n$ region from the image and replace the original pixel values with their mean. The loss function during the second stage of training is:
\begin{equation}
\label{loss2}
\mathcal{L}_R=\mathcal{L}_V+\lambda_{J}\mathcal{L}_{J},
\end{equation}
where $\lambda_{J}$ is hyper-parameter. $\mathcal{L}_V$ is the mean squared error of voxel occupancy values:
\begin{equation}
\label{lossV}
\mathcal{L}_V=\frac{1}{N_v} \sum_{i=1}^{N_v} |f(V) - f^*(V)|^2,
\end{equation}
where $f^*(V)$ is the ground truth. 
The positions of human body joints $J_{3D}\in \mathbb{R}^{24\times 3}$ can be obtained by processing the joint features $\mathcal{F}_j$ through a Multilayer Perceptron (MLP). Therefore, the 3D joints loss can be formulated as follows:
\begin{equation}
\label{lossJ}
\mathcal{L}_J=\frac{1}{N_J} \sum_{i=1}^{N_J} |J_{3D} - J^*_{3D}|, 
\end{equation}
where $J^*_{3D}$ is the ground-truth joint position.

\begin{table*}[!h]
\centering
\caption{Quantitative comparisons with the state-of-the-art methods on the Adobe and MonoPerfCap datasets. The quantitative results of our method, combining the Temporally Consistent Reconstruction module (TM) and Mesh Refinement module (MM), are also presented. The best and second-best results are marked in \textbf{bold} and \underline{underlined}, respectively.} 
\begin{tabular}{lcccccc}
\toprule
\multirow{2}{*}{Methods} & \multicolumn{3}{c}{Adobe\_P2}                        & \multicolumn{3}{c}{MonoPerfCap} \\ \cmidrule(r){2-4} \cmidrule(r){5-7}
     & P2S $\downarrow$ & Chamfer $\downarrow$  & Normal $\downarrow$ & P2S $\downarrow$ & Chamfer $\downarrow$  & Normal $\downarrow$ \\ \midrule
PIFu \cite{PIFu}     & 3.386   & 2.916 & 0.192   & 3.467 & 2.812 & 0.173 \\
PIFuHD \cite{PIFuHD} & 3.268   & 2.856 & 0.187   & 3.201 & 2.673 & 0.169 \\
PaMIR \cite{PaMIR}   & 2.193   & 2.126 & 0.169   &1.878 &1.695 &0.129 \\
ICON \cite{ICON}    & 2.266    & 2.134 & 0.166   &1.825 &1.626 &0.127 \\
ECON \cite{ECON}   & 2.215     & 2.141 & 0.163   &1.798 &1.619 &0.124 \\
FOF \cite{FOF} & 2.278 & 2.112& 0.155 & 1.825 & 1.526& 0.127\\
TCR \cite{TCR}     & \underline{1.924}     &  1.922 & \underline{0.115}  &\underline{1.563} &\underline{1.492} &\underline{0.104} \\
\midrule
ours               & 1.975     &  \underline{1.916} & 0.126  &1.726 &1.586 &0.110 \\ 
Ours+TM+MM & \bf{1.873} & \bf{1.811}& \bf{0.106} &\bf{1.552} &\bf{1.367} &\bf{0.096} \\
 \bottomrule
\end{tabular}
\label{tab:qualitative}
\end{table*}

\subsection{Implementation details}
\label{MeshR}

\noindent {\bf{Optimization.}} More accurate SMPL estimation can provide better human priors for implicit reconstruction. Therefore, we adopted the optimization strategy of ICON \cite{ICON}, utilizing the difference between the predicted normal map and the rendered SMPL normal map to optimize the model's shape, pose and translation parameters ($\beta, \theta, t$). The cost function is: 
\begin{equation}
\label{cost}
\mathcal{L}_P=\min_{\beta, \theta, t}(\mathcal{L}_S+\lambda _{N}\mathcal{L}_N),
\end{equation}
Where $\mathcal{L}_S$ is L1 loss between the SMPL normal map and human mask. $\mathcal{L}_N$ is the loss of normal maps and $\lambda _{N}$ is the weight coefficient. 

\noindent {\bf{Post-processing.}} To compare with the previous work, TCR \cite{TCR}, and prove the superiority of this method, we incorporate the Temporally Consistent Reconstruction module (TM) and Mesh Refinement module (MM) from TCR work as a post-optimization step. The first module enhances the completeness of the reconstructed human body by extracting temporally consistent shape information, while the second module refines the surface details by adjusting vertex positions using normal orthogonal constraints. We only used post-optimization steps during quantitative comparisons to demonstrate that incorporating optimization steps in this method can achieve state-of-the-art performance. All qualitative presentations have not undergone post-optimization steps, in order to more visually show the improvement of our method in reconstruction quality of single frame.

\noindent {\bf{Transformer Models.}} The encoder structures of the spatial transformer and temporal transformer are both based on the visual transformer (ViT) \cite{VIT}, but they have different numbers of encoder layers $l$, hidden size $h$ and heads $m$. The specific settings will be introduced in the Sec. \ref{tran_size}. The decoder structure of the temporal transformer is based on the segmenter \cite{Segmenter}. To guide the network's learning of the correspondence between different parts of the human body, we introduced joint tokens into the decoder. The joint token has a dimension of 14.  For the temporal encoder, each clip contains 7 frames. We use copy operation to complete the absent video frames at the beginning and end.

\section{Experiments}
\label{Experiments}

This section shows the performance of our method and the comparison with the state-of-the-art methods. All experiments are executed on an Intel Xeon Gold 6226R 2.9 GHz CPU and an NVIDIA GeForce RTX 3090 GPU. Our method uses a weak-perspective camera model.

\begin{table*}[!h]
\centering
\caption{Performance comparison of different Transformer model variants. Three different encoder parameter configurations were tested for both the spatial transformer and temporal transformer. For the spatial transformer, we employ L2 loss of the back normal map (L-NB) as the performance evaluation metric. As for the temporal transformer, we report the Point-to-surface Euclidean Distance (P2S) of the reconstructed model.} 
\begin{tabular}{cccccccc}
\toprule
 & Backbone & Layers $l$ & Hidden size $h$ & Heads $m$ & Params  &L-NB & P2S \\
\midrule
\multirow{3}{*}{S-Trans}         & ViT-B  & 12 & 768 & 12  & 86M  & 0.195 & -\\ 
                                 & ViT-L & 24  & 1024 & 16 & 307M & 0.178 & - \\ 
                                 & ViT-H & 32  & 1280 & 16 & 632M & 0.172 & - \\ 
\midrule
\multirow{3}{*}{T-Trans}         & ViT-B  & 12 & 768 & 12  & 86M  & - & 2.212\\ 
                                 & ViT-L & 24  & 1024 & 16 & 307M & - & 2.095\\ 
                                 & ViT-H & 32  & 1280 & 16 & 632M & - & 1.975\\ 
 \bottomrule
\end{tabular}
\label{tab:Trans_size}
\end{table*}

\subsection{Dataset}
In the subsequent experiments, we use the Adobe dataset \cite{Adobe} as the training set for the entire network and THuman2.0 \cite{THuman} as the training dataset for the normal prediction network. Meanwhile, the MonoPerfCap dataset \cite{MonoPerfCap} is used as the test data. To ensure a fair comparison, all the compared methods are trained on the Adobe dataset.

\noindent {\bf{Adobe:}} The Adobe dataset comprises 10 video sequences of three subjects. All video sequences are captured by eight cameras in an indoor environment with green screen. We extract 100 frames from each sequence as training data named Adobe\_P1, and the remaining frames are used as test data named Adobe\_P2.

\noindent {\bf{Thuman2.0:}} THuman2.0 contains 500 scan models captured by a dense DSLR rig, with corresponding SMPL(-X) \cite{SMPL,SMPL-X} parameters for each model. We use this dataset as the training dataset for spatial transformer. The input images are generated by a renderer. The normal map is obtained by mapping the normals of the 3D model to the image space.

\noindent {\bf{MonoPerfCap:}} The MonoPerfCap dataset comprises both outdoor and indoor data. We only utilize 7 outdoor videos captured from 7 subjects as supplementary data for evaluation of generalization. Meanwhile, we employ two videos with ground truth \cite{truth1,truth2} for quantitative evaluation.

\subsection{Evaluation Metrics}
We assess the quality of human body surface reconstruction through the utilization of three conventional metrics: Chamfer Distance (Chamfer), Point-to-surface Euclidean Distance (P2S), and Normal Reprojection Error (Normal).

\noindent {\bf{Chamfer:}} The Chamfer Distance is defined as the sum of distances between the nearest points between the point clouds sampled from the reconstructed human surface and those sampled from the original scanned data. A smaller Chamfer Distance indicates a higher global similarity between the reconstructed human surface and the original scanned data.

\noindent {\bf{P2S:}} The Point-to-surface Euclidean Distance is the distance between the vertices on the reconstructed human surface and their closest points on the original scanned data surface. This metric is used to measure the local accuracy and fidelity of the reconstruction. A smaller P2S reflects higher surface precision in the reconstruction.

\noindent {\bf{Normal:}} The Normal Reprojection Error refers to the L2 error between the normal vectors on the reconstructed 3D surface and those on the original scanned data surface. A smaller L2-norm shows higher reconstruction accuracy.

\begin{figure*}[!t]
\centering
\includegraphics[width=\linewidth]{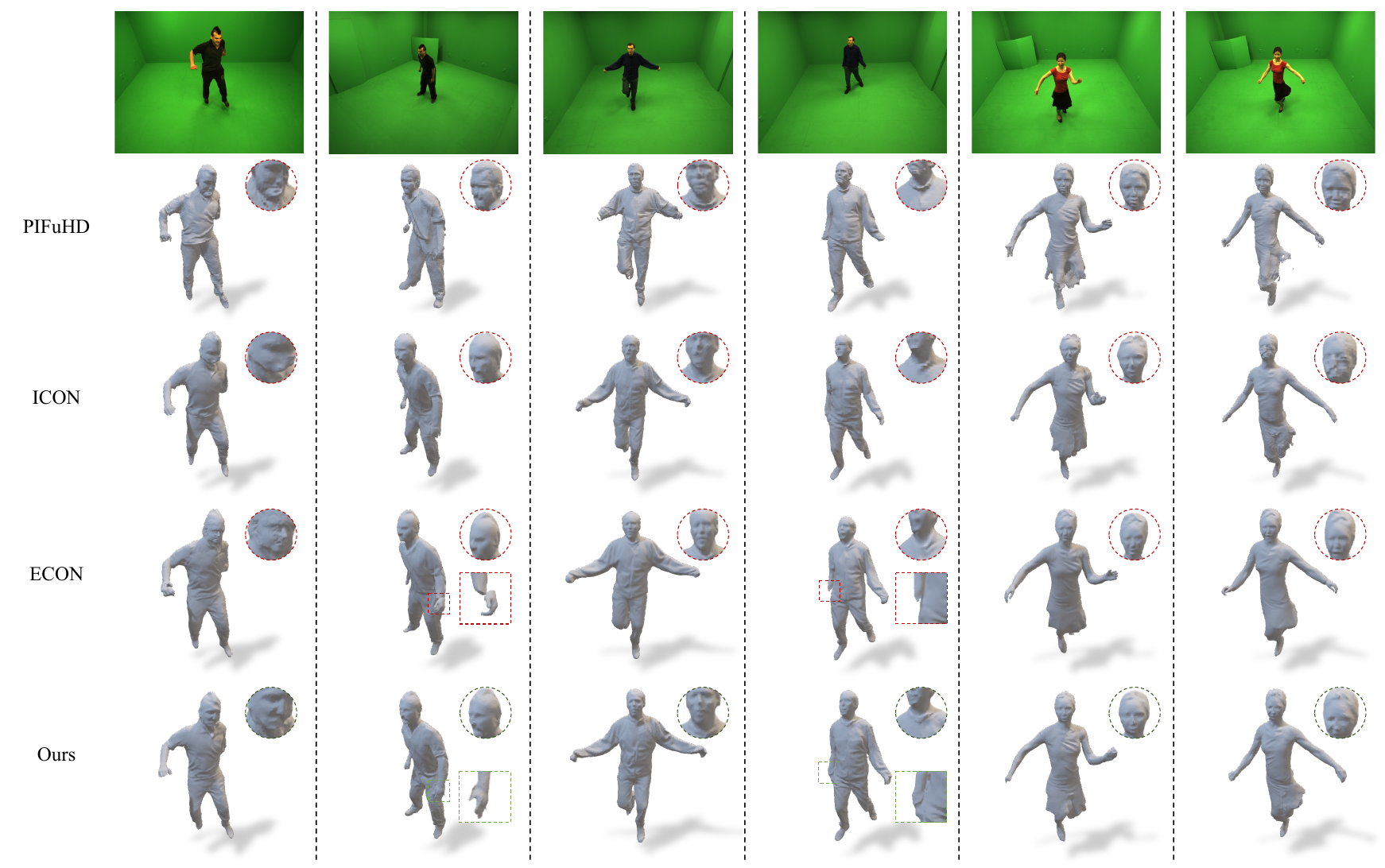}
\caption{Qualitative comparison with the state-of-the-art method on Adobe dataset \cite{Adobe}. The improved blurry image regions are outlined with a dashed line. Our method demonstrates improvements over PIFuHD \cite{PIFuHD} and ICON \cite{ICON} in the reconstruction of details in blurry regions of the image and in preserving the structural integrity of the human body. ECON \cite{ECON} can also maintain the integrity of local structures, but its performance in details is inferior to our method. Additionally, as indicated by the boxed area, it is vulnerable to the accuracy of parametric model estimation.}
\label{qualitative_indoor}
\end{figure*}

\begin{table}[!h]
\centering
\caption{Quantitative comparison of front and back normal L2 losses (L-NF \& L-NB)  and reconstruction errors between the CNN-based normal estimation network (Conv), the normal estimation network with replaced inputs (Conv-refine), and our spatial transformer (Trans) on the Adobe dataset.} 
\begin{tabular}{cccccc}
\toprule
\multicolumn{1}{l}{} & \multicolumn{2}{c}{Normal}                        & \multicolumn{3}{c}{Reconstruction}           \\ \cmidrule(r){2-3} \cmidrule(r){4-6}
                                & L-NF $\downarrow$       & L-NB $\downarrow$       & P2S $\downarrow$           & Chamfer $\downarrow$       & Normal $\downarrow$       \\ \midrule
Conv   & 0.155          & 0.213          & 2.266    & 2.134 & 0.166          \\
Conv-refine     & 0.151          & 0.192          & 2.198     &  2.083 & 0.161   \\ 
Trans  & \textbf{0.143} & \textbf{0.178} & \textbf{2.123} & \textbf{2.023} & \textbf{0.153} \\ \bottomrule
\end{tabular}
\label{tab:normal_pre}
\end{table}

\subsection{Transformer Parameters}
\label{tran_size}
We investigated the impact of parameter variations on reconstruction quality to determine the encoder parameters for the spatial transformer (S-Trans) and temporal transformer (T-Trans). Three configurations are set, namely ViT-B, ViT-L and ViT-H, by adjusting the number of encoder layers $l$, hidden size $h$, and the number of heads $m$. For the spatial transformer, we showcased the prediction errors of the back normal map to reflect the impact of parameter variations, while for the temporal transformer, we documented the P2S metric of the reconstruction model. The results are presented in Table \ref{tab:Trans_size}. It can be observed from the table that ViT-H shows only marginal improvement in normal estimation compared to ViT-L, despite doubling the parameter count. In consideration of both reconstruction quality and resource requirements, the spatial encoder used ViT-L. Simultaneously, due to the excessive computational cost associated with further increasing the model size, the encoder of the temporal transformer adopts the configuration of ViT-H.

\subsection{Comparison with the State-of-The-Art}
To demonstrate the performance advantages of our method compared to the state-of-the-art works, we conducted qualitative and quantitative comparisons. We evaluated our method using Chamfer, P2S, and L2-norm metrics on Adobe\_P2 and MonoPerfCap datasets. Additionally, we conduct a separate evaluation to assess the quality of the normal maps predicted by the spatial transformer.

\noindent {\bf{Overall Performance.}} A spatial transformer and a temporal transformer are employed to enhance the quality of reconstructing human models in our method. In order to demonstrate the performance improvement of STT, we present the quantitative evaluation results of our method on Adobe\_P2 and MonoPerfCap datasets, and compare it with state-of-the-art methods \cite{PIFu,PIFuHD,PaMIR,ICON,ECON,TCR} in Table \ref{tab:qualitative}. As shown in the table, our method outperforms the earlier methods across all three metrics on both datasets. However, our method performs worse than TCR, which has two post-optimization modules: Temporally Consistent Reconstruction Module (TM) and Mesh Refinement Module (MM). By incorporating these post-optimization modules into our method as a post-optimization step, our method achieves the best results. In Fig. \ref{qualitative_indoor}, we qualitatively show several instances from Adobe\_P2. Although the indoor lighting is fixed, due to human movement, local areas of the human body may be occluded by light or affected by fast motion, resulting in ambiguity. We have highlighted the improved regions within the frame. As can be seen in the figure, our method achieves a certain improvement in these local areas, especially in areas with deep colors and affected lighting, such as hair and dark clothing.

\begin{figure*}[!t]
\centering
\includegraphics[width=\linewidth]{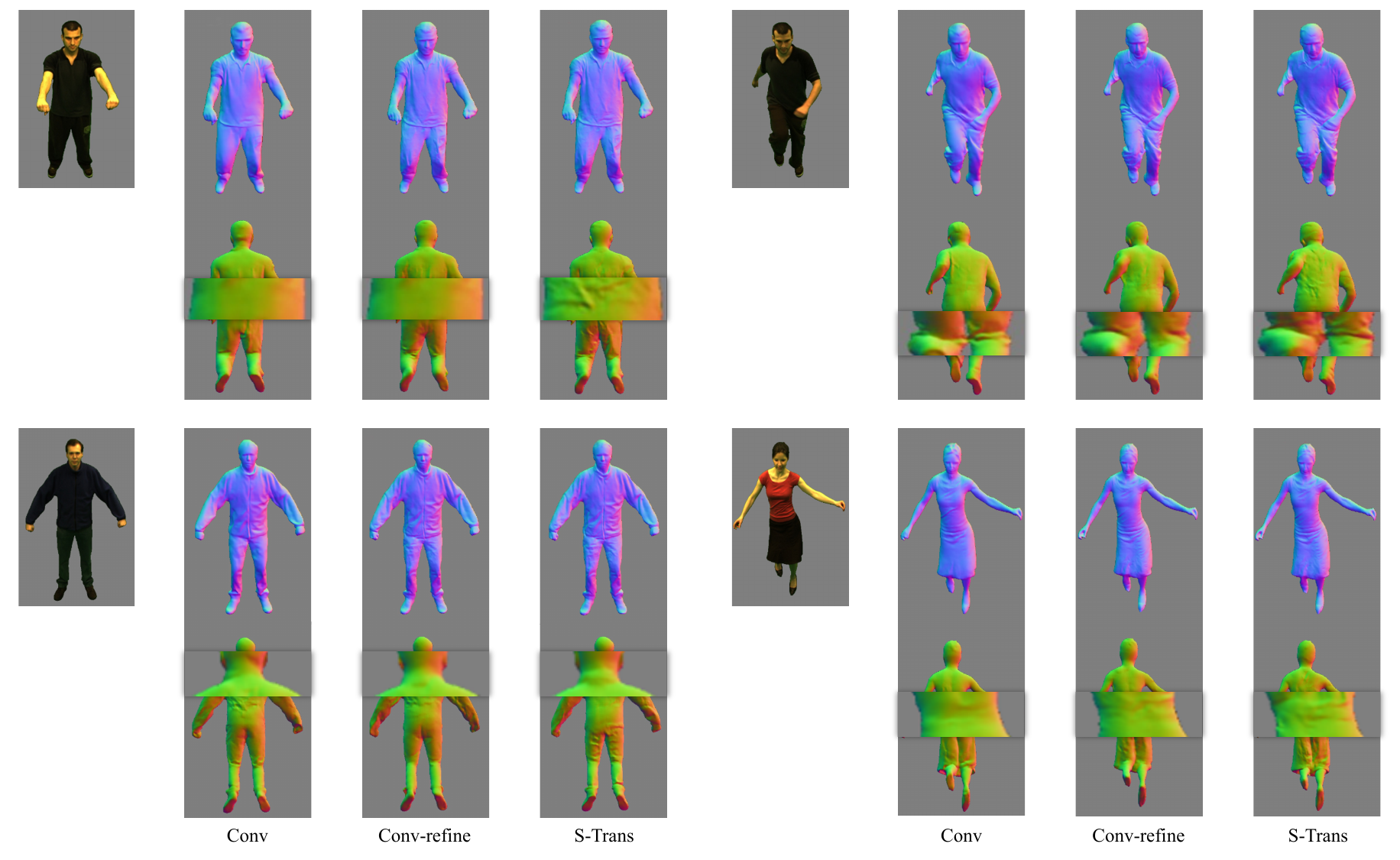}

\caption{Qualitative comparison of the spatial transformer (S-Trans) for normal prediction with the 2D convolutional module from ICON \cite{ICON} (Conv) and the 2D convolutional module from TCR \cite{TCR} with replaced inputs (Conv-refine). When compared to convolutional networks, S-Trans's extraction of global information greatly enhances the prediction of details in invisible areas.}
\label{fig:normal_pre}
\end{figure*}

\noindent {\bf{Normal Maps.}}
The convolutional network has difficulty establishing global correlations for back normal prediction. Therefore, A spatial transformer is employed to establish global correlations aiming to enhance the accuracy of back normal prediction. To demonstrate the efficacy of this module, we conducted an independent evaluation of normal map prediction. Table \ref{tab:normal_pre} compares the performance of the convolutional module used for normal prediction in ICON (Conv), the modified module with replaced inputs from TCR (Conv-refine), and our spatial transformer (Trans). We computed the L2 loss between the predicted and ground truth normal maps. Furthermore, we assessed the impact of the three normal estimation modules on the reconstruction quality by replacing the normal module in ICON. The results indicate that our spatial transformer outperforms the previous normal prediction modules. Additionally, Fig. \ref{fig:normal_pre} illustrates several examples of predicted normal maps, demonstrating that the spatial transformer improves the quality of both front and back normal maps. Notably, due to the establishment of global correlations, our approach provides significantly richer detailed information in back normal prediction.

\begin{table}[!h]
\centering
\caption{Ablation Study. Three sets of ablation experiments were conducted, involving ``without spatial transformer'', ``without temporal transformer'', and ``without Joint Tokens''. } 
\begin{tabular}{cccc}
\toprule
& P2S $\downarrow$ & Chamfer $\downarrow$  & Normal $\downarrow$ \\ \midrule
w/o S-Trans        & 2.054    & 1.998 & 0.141 \\   
w/o T-Trans        & 2.123    & 2.023   &   0.153  \\
w/o J-Tokens        & 2.213    & 2.101   &   0.159   \\   
\bottomrule
\end{tabular}
\label{tab:ablation}
\end{table}

\begin{figure*}[!t]
\centering
\includegraphics[width=\linewidth]{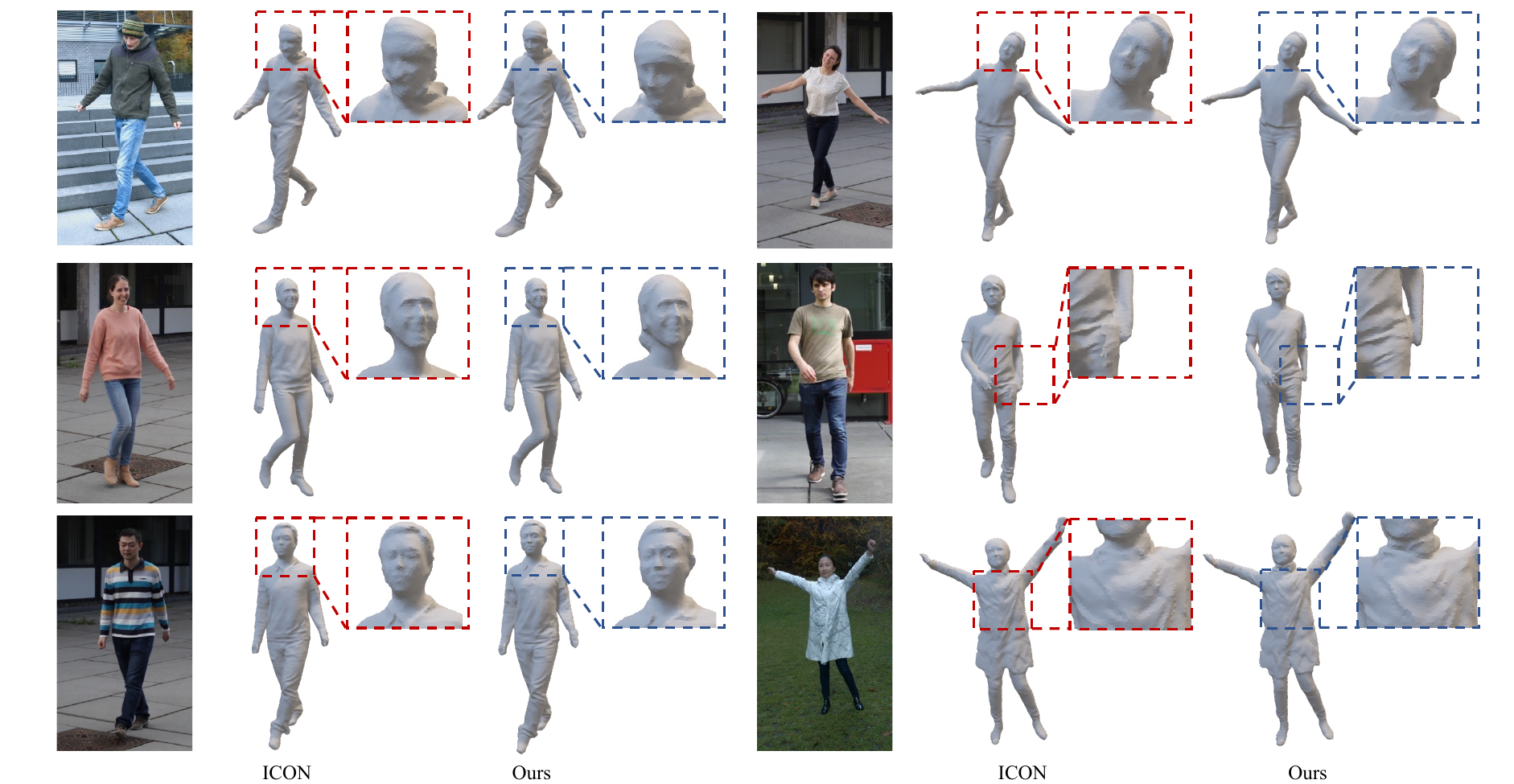}

\caption{The results of our method on outdoor dataset MonoPerfCap \cite{MonoPerfCap} and the comparison with the baseline ICON \cite{ICON} are presented. In outdoor scenarios, characterized by intricate and fluctuating lighting conditions, our approach has demonstrated commendable generalization even under these circumstances.}
\label{fig:quali_out}
\end{figure*}

\noindent {\bf{Generalization.}}
The temporal transformer improves reconstruction accuracy by leveraging temporal information, mitigating errors arising from local ambiguity. Consequently, our method exhibits superior generalization capabilities, remaining highly robust even in outdoor conditions with significant variations in lighting and a larger range of activities. To illustrate the generalization capability of our method, we showcase its reconstruction performance on outdoor data from MonoPerfCap \cite{MonoPerfCap} in Fig. \ref{fig:quali_out}. Outdoor image captures exhibit diverse and complex lighting conditions, leading to degraded reconstruction quality in the presence of suboptimal illumination. We have outlined the significantly improved areas with dashed lines. From the figure, it is evident that our method outperforms the baseline ICON \cite{ICON} in the reconstruction quality of outdoor data, particularly in locally darker regions.

\subsection{Ablation Study}
Our method utilizes a spatial transformer to enhance the quality of normal prediction and employs the spatial transformer to enhance predictions in ambiguous regions of the image. To demonstrate the roles of both the spatial transformer and temporal transformer in the reconstruction process, we conducted two ablation experiments, ensuring a credible evaluation of their respective contributions. One set involved replacing the spatial transformer with a normal estimation network from ICON (w/o S-Trans), while the other set omitted the temporal features extracted by the temporal transformer (w/o T-Trans). The results presented in Table \ref{tab:ablation} illustrate that both modules contribute to the improvement in reconstruction quality. Although there is an increase in the predicted details of the back normal map, it struggles to entirely match the ground truth. Hence, the temporal transformer exhibits a more discernible impact on the overall enhancement of reconstruction quality. In Sec. \ref{TemporalT}, joint tokens are introduced to guide the temporal transformer in capturing correspondences between frames and extracting temporal features. To demonstrate the role of joint tokens, we also evaluated the spatial temporal transformer without joint tokens (w/o J-Tokens). Compared with the third row in Table \ref{tab:ablation} (w/o T-Trans), the temporal features extracted by the spatial temporal transformer without joint tokens have a negative impact on the reconstruction quality. This indicates that Joint Tokens play a crucial role in ensuring that the spatial temporal transformer accurately extracts temporal features.

\begin{figure}[!t]
\centering
\includegraphics[width=\linewidth]{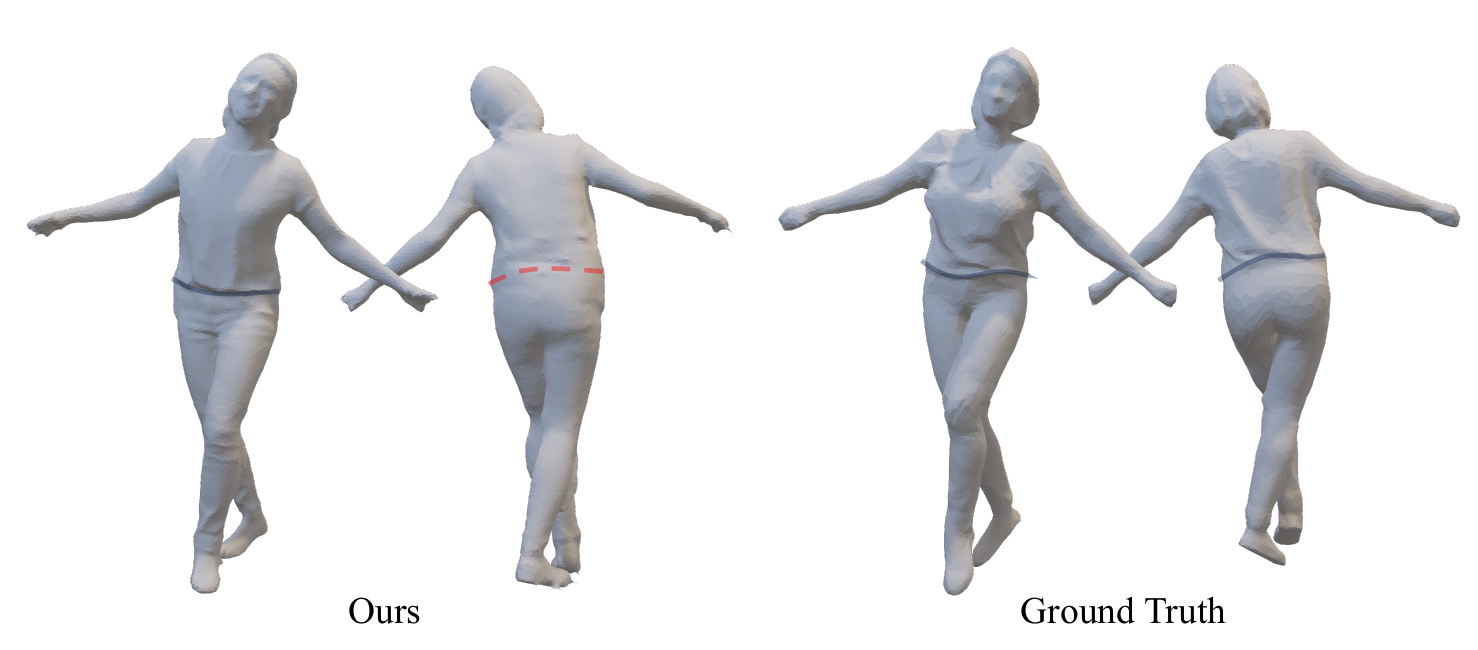}
\caption{The delineation of garment contours holds significant importance in the sensory evaluation of reconstructed human model. Compared to the multitude of possibilities in the detail of the back body, the back contour is relatively fixed due to determinate clothing types. However, as indicated by the red line, the reconstructed results in our method may sometimes lack the contour line of clothing.}
\label{fig:limitation}
\end{figure}

\subsection{Limitations}

Our method addresses the issue of back detail ambiguity and human image ambiguity in clothed human reconstruction through the spatial-temporal transformer. Nevertheless, our method has two inherent limitations. Firstly, as shown in Fig. \ref{fig:limitation}, the reconstruction of the back sometimes lacks the necessary contour lines of clothing. There are multiple solutions for back details, but clothing contours is relatively certain due to determinate clothing types. Therefore, generating complete clothing contours is what we need and more reasonable. Secondly, the amelioration of local ambiguity is contingent on the improvement of image quality in neighboring frames. In situations where ambiguity persists due to lighting conditions, significant improvement is difficult. 

\section{Conclusions and discussion}

In this paper, we propose a novel approach for 3D clothed human reconstruction, termed the Spatial Temporal Transformer (STT), using data captured from a monocular camera. To enhance the prediction of back normal maps, a spatial transformer is introduced to establish global correlations within the texture. Simultaneously, we propose a temporal transformer to extract temporal features, mitigating the impact of local ambiguity caused by variations in lighting and human motion. Experimental results demonstrate that our method enhances the details in back normal map predictions and mitigates the effects of local ambiguity, thereby improving the generalization of the method. However, the reconstructed back details from our method sometimes lack the necessary clothing contours. This problem may be considered in our future works.

\newpage

 




\vfill

\end{document}